\documentclass[draft]{agujournal2019}
\usepackage{url} %
\usepackage{lineno}
\usepackage{nicematrix}
\usepackage[inline]{trackchanges} %
\usepackage{soul}
\usepackage{amsfonts}
\usepackage{amsmath}
\usepackage{subcaption}

\draftfalse

\journalname{Journal of Advances in Modeling Earth Systems (JAMES)}

\begin{document}

\title{Learning Prognostic Variables for AI Convective Parameterizations via Symbolic Distillation}

\authors{Jurij Sch\"onfeld\affil{1, 2}, Tom Beucler\affil{3, 4}, Julien Savre\affil{1}, Steven Sherwood\affil{5, 6}, Veronika Eyring\affil{1, 2}}

\affiliation{1}{Deutsches Zentrum für Luft- und Raumfahrt, Institut für Physik der Atmosphäre, Oberpfaffenhofen, Germany}
\affiliation{2}{University of Bremen, Institute of Environmental Physics (IUP), Bremen, Germany}
\affiliation{3}{Faculty of Geosciences and Environment, University of Lausanne, Lausanne, Switzerland}
\affiliation{4}{Expertise Center for Climate Extremes, University of Lausanne, Lausanne, Switzerland}
\affiliation{5}{Climate Change Research Centre, University of New South Wales, Sydney, New South Wales, Australia}
\affiliation{6}{ARC Centre of Excellence for 21st Century Weather, University of New South Wales, Sydney, New South Wales, Australia}

\correspondingauthor{Jurij Schönfeld}{jurij.schoenfeld@dlr.de}

\begin{keypoints}
\item An autoencoder compresses past, coarse state observations into low-dimensional memory variables for subgrid parameterization
\item In L96 and km-scale simulations, symbolic distillation yields Markovian equations to prognose memory variables forced by present observables
\item Prognostic memory models outperform present-state-only parameterizations, improving the diurnal cycle of tropical land precipitation
\end{keypoints}

\begin{abstract}
Hybrid AI-physics climate modeling aims to improve coarse ($\sim$100km-resolution) Earth system models by learning to parameterize subgrid processes from high-fidelity data. However, this so far mostly involves local-in-time, diagnostic parameterizations, in which the subgrid state depends only on the current coarse state with no memory of previous states, which is unrealistic for processes such as convection that have intrinsic persistence. To address this, we enhance local-in-time parameterizations by learning prognostic variables that compactly carry important, additional past information where no explicit sub-grid information is available. First we compress past information into a low-dimensional latent space using an autoencoder, which then informs a neural network trained to parameterize targeted subgrid-scale processes. We then replace the autoencoder with symbolic equations that govern the time evolution of the latent variables, yielding additional prognostic memory variables that can be integrated alongside the resolved atmospheric state. We evaluate this approach on two systems: the Lorenz-96 model (online) and surface precipitation from high-resolution atmospheric simulations (offline). A forced multivariate linear ordinary differential equation recovers most of the added value achieved by the autoencoder-based approach in both experiments. Benchmarked against diagnostic parameterizations without memory, our memory-informed approach improves climate statistics and temporal structure, including a realistic diurnal cycle of tropical land precipitation. 
\end{abstract}

\section*{Plain Language Summary}
Earth system models (ESMs) evolve the Earth's climate through a set of governing equations on a coarse resolution grid to maintain computational feasibility. Physical processes below the computed resolution need to be approximated with parameterizations. Atmospheric patterns, like organized moisture patches, leave a fingerprint on the time series of grid-scale state variables inducing a memory of the past. We improve parameterizations by leveraging this information with a Machine Learning (ML) model creating a new set of variables that expand the governing equations of the dynamical core. The ML model can be replaced by approximating its behavior with a set of evolution equations that can be interpreted and simulated efficiently unlike the original ML model. We test our approach with an atmospheric toy model and a realistic parameterization of precipitation. We find that the new equations recover most of the predictive skill added by ML unlike parameterizations without memory.

\section{Introduction}
Earth system models (ESMs) provide the primary computational framework for understanding past climate variability and projecting future climate change. Their importance continues to grow as anthropogenic greenhouse gas emissions rise and global temperatures approach or exceed the targets established under the Paris Agreement, with implications for mitigation, adaptation, and climate risk assessment \cite{jones_national_2023, ipcc_impacts_2022}. %
Over successive CMIP generations, ESMs have demonstrated substantial progress in reproducing large-scale climate statistics, including near-surface temperature, precipitation, and top-of-atmosphere radiation fields \cite{bock_quantifying_2020, carvalho_how_2022}. At the same time, persistent systematic biases remain, particularly in the representation of clouds, convection, and tropical circulation patterns such as the Intertropical Convergence Zone \cite{tian_double-itcz_2020}. Cloud feedbacks remain the major source of uncertainty in climate sensitivity and future warming projections \cite{zelinka_causes_2020}, even when overall uncertainty is reduced by incorporating understanding of feedback processes, historical climate records, and paleoclimate \cite{sherwood_assessment_2020}. 

For the physical components of the climate system, the governing equations %
must be approximated numerically on discrete computational grids. The computational cost of resolving all relevant scales explicitly is prohibitive for centennial climate integrations, forcing operational climate models to employ horizontal resolutions on the order of hundreds of kilometers. As a result, many important processes—including moist convection, turbulence, cloud microphysics, and radiative interactions—remain unresolved and must be represented through parameterizations. These parameterizations constitute a major source of model uncertainty because they approximate the aggregate influence of unresolved subgrid-scale dynamics on the resolved large-scale flow.

Machine learning (ML) parameterizations have emerged as a promising alternative to conventional heuristic or semi-empirical schemes. In hybrid ML-enhanced Earth system modeling approaches, the resolved large-scale dynamics continue to be governed by the numerical solution of physical equations, while unresolved subgrid-scale tendencies are estimated from data-driven models trained on high-resolution simulations or observations \cite{eyring_pushing_2024, eyring_ai-empowered_2024}. ML parameterizations can reproduce complex subgrid-scale processes with high accuracy and, in some cases, remain stable over several decades when coupled online to dynamical models \cite{yuval_use_2021, hu_stable_2025, heuer_beyond_2026}. However, major challenges remain, including limited interpretability \cite{brenowitz_interpreting_2020}, computational cost \cite{han_decadal_2025}, physical inconsistency \cite{beucler_enforcing_2021}, and poor generalization outside the training distribution \cite{beucler_climate-invariant_2024}.

A frequent limitation of both conventional and ML parameterizations is their diagnostic nature, as they compute subgrid-scale contributions directly from instantaneous prognostic variables provided by the dynamical core. This limitation is made explicit by the Mori–Zwanzig formalism, which shows that eliminating unresolved variables from a dynamical system generally produces non-Markovian evolution equations containing explicit memory terms \cite{lucarini_theoretical_2023}. In the case of convection, the convective quasi-equilibrium (CQE) assumption provides the constraint to diagnose the closure variables by assuming that the resolved dynamics slowly destabilize the atmosphere, while convection responds rapidly to counteract this destabilization \cite{arakawa_interaction_1974}. Although CQE has provided a powerful foundation for many traditional convective parameterizations, it fails to describe situations in which the convective response to large-scale forcing is not instantaneous.
Indeed, convection and cloud organization exhibit pronounced temporal memory and mesoscale organization, suggesting that unresolved dynamics may depend strongly on prior system evolution \cite{tobin_does_2013,colin_identifying_2019,hwong_assessing_2023}. This memory effect is important because it governs the timing, intensity, and spatial clustering of convective activity, which in turn modulates large-scale circulation, moisture transport, and climate feedbacks over timescales of hours to days. Physically, memory is encoded through high-resolution atmospheric patterns, particularly subgrid moisture heterogeneities (moist and dry patches) and cold pool dynamics \cite{muller_spontaneous_2022}. In atmospheric convection, memory effects are physically linked to evolving organization states that appear as temporal signatures in large-scale observables \cite{colin_identifying_2019}. 

To challenge CQE by incorporating temporal memory into convection parameterizations, traditional approaches typically introduce prognostic memory variables grounded in distinct physical mechanisms: for example, modulating deep convection entrainment based on recent surface precipitation \cite{lock_performance_2024}, or explicitly tracking cold-pool evolution \cite{grandpeix_density_2010, rooney_c-pool_2022}. However, it remains unclear which physical formulation best represents convective memory. In contrast, ML parameterizations circumvent the need to prescribe explicit physical memory by learning temporal dependencies directly from data. By processing sequences of past states, temporal ML architectures based on recurrent neural networks \cite{song_physically_2025} and transformers \cite{wang_climt-paraformer_2026} can implicitly capture convective history and environmental preconditioning. At the cost of physical interpretability and increased model complexity, data-driven approaches including past coarse states have improved online stability, precipitation statistics, and generalization under climate change conditions compared to instantaneous ML parameterizations \cite{han_ensemble_2023, lin_navigating_2025}. Complementing these temporal approaches, latent-space representations derived from high-resolution simulations have shown that compressed representations of unresolved convective organization can substantially improve instantaneous coarse-scale precipitation predictions \cite{shamekh_implicit_2023}. Because high-resolution information is not available in the parameterization scenario, the authors provide a simple estimate of how well the latent space variables can be predicted from the previous time step. However, how to model the full temporal evolution of such latent variables while controlling error accumulation remains unresolved. %
To serve as prognostic state variables, these latent variables must remain stable and retain predictive information over long rollouts, requiring carefully constructed evolution equations.

The evolution of latent variables has been studied extensively in reduced-order modeling of nonlinear dynamical systems \cite{bonneville_comprehensive_2024}. Applications combine deep autoencoder architectures with system identification to discover low-dimensional latent coordinates together with explicit evolution equations governing their dynamics \cite{champion_data-driven_2019} or system identification from partial measurements via time-delayed encoder embeddings \cite{bakarji_discovering_2023}. The special case where latent dynamics are enforced to be linear connects to Koopman theory with promising applications, such as the data-driven development of moment-based microphysics schemes \cite{lamb_reduced-order_2024}, and the interpretable embedding \cite{lusch_deep_2018} and forecasting \cite{nayak_temporally_2025} of nonlinear dynamical systems.

In this work, we investigate how temporal memory carried by unresolved subgrid-scale processes can be learned to improve ML parameterizations for ESMs. Our framework builds upon the organization-informed parameterization introduced by \cite{shamekh_implicit_2023} without requiring high-resolution input. First, an autoencoder compresses past observations into a low-dimensional latent representation capturing unresolved dynamical information. Second, symbolic distillation identifies interpretable Markovian evolution equations for these latent variables forced by present-state, coarse-scale observables. The resulting framework yields a set of data-driven prognostic equations that represent the evolution of latent variables encoding memory effects and can be integrated in time alongside the resolved model state.

We evaluate the proposed approach in two settings: the Lorenz-96 system, as an idealized testbed for multiscale dynamics; and a kilometer-scale atmospheric simulation for coarse-grained precipitation parameterization, to compare with \citeA{shamekh_implicit_2023} who used similar data. The remainder of this paper is organized as follows. Section~2 introduces our memory-informed framework and details the two application cases. Sections~\ref{sec:L96} and \ref{sec:DYAMOND_precip} present the results of our two experiments. Section~\ref{sec:conclusion} continues the discussion of our findings and their implications for ESM development, and outlines directions for future work.

\section{Methods}
\subsection{Parameterization Architecture: A Deep Memory Autoencoder}
\begin{figure}
    \centering
    \includegraphics[width=0.90\linewidth]{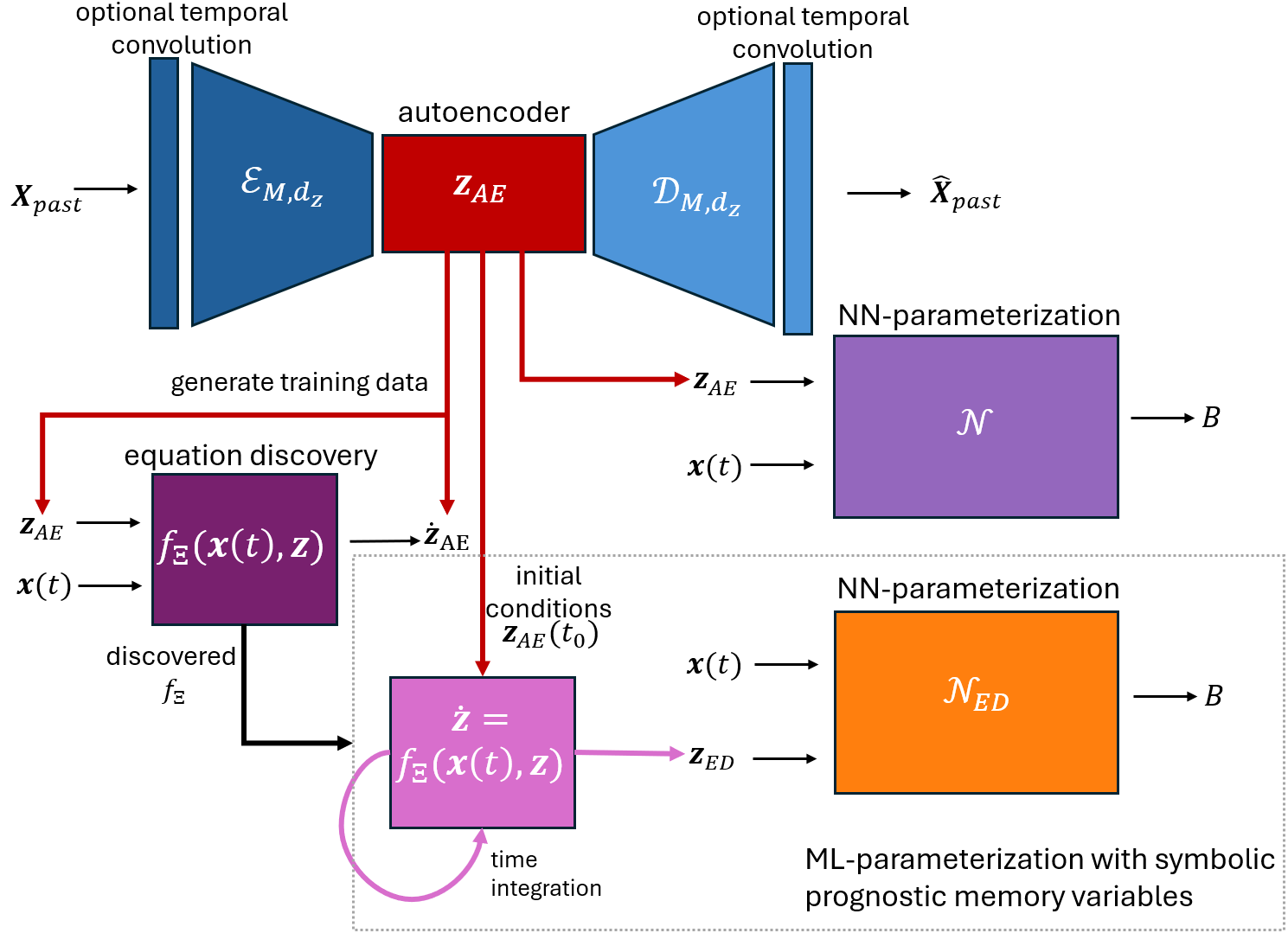}
    \caption{%
    Two prognostic parameterizations are compared. A parent model $\mathcal N$ that generates latent space variables from past observables using an autoencoder $\mathbf{z}_{AE}$ and a distilled model $\mathcal N_{ED}$ that replaces the autoencoder with an ODE derived via equation discovery (ED) forming a different set of prognostic variables $\mathbf{z}_{ED}$.}
    \label{fig:architecture}
\end{figure}
The general approach for parameterization assumes that the system's state variables can be decomposed into a multi-level system with clear scale separation. One part of the state $\mathbf{x}\in\mathbb R^{d_x}$ evolves slowly and is represented by coarse-scale variables evolved by a dynamical core. The other part of the state, $\mathbf{y}\in\mathbb R^{d_y}$, corresponds to the unresolved sub-grid states. In general, $\mathbf{x}$ and $\mathbf{y}$ are coupled bi-directionally, and the parameterization challenge is described by finding a %
functional relation between a set of input features $\mathbf{X}_{\text{I}}$---which must be obtained from the coarse-scale state variables---to the sub-grid scale contribution $B$. It represents the impact of the unresolved variables $\mathbf{y}$ onto the evolution of the resolved state variables. %
ML-based parameterizations approximate $B$ using neural network (NN) architectures, which offer high representational flexibility but require substantial training data to effectively optimize their numerous parameters.

Incorporating memory effects in ML-based parameterizations presents additional challenges. The current paradigm adopted by \citeA{lin_navigating_2025}, \citeA{han_ensemble_2023}, \citeA{heuer_beyond_2026}, or \citeA{behrens_simulating_2025} to capture convective memory relies on adding past atmospheric states or past subgrid contribution predictions to the input vector
\begin{equation}
    \mathbf{X}_I=\{\mathbf{x}(t), \underbrace{\mathbf{x'}(t-\Delta t), ..., \mathbf{x'}(t-M\Delta t)}_{\mathbf{X}_{\text{past}}}\}
\end{equation}
covering a time span $M\Delta t$, with $\Delta t$ the host model time step. The input variables $\mathbf x'\in\mathbb R^{d_{x'}}$, used for the memory representation, are a subset of past time steps from the present time input variables $\mathbf{x}(t)\in\mathbb R^{d_x}$.

Here we adopt a different approach, whereby an autoencoder is employed to isolate memory effects encoded at the subgrid scales. In particular, the autoencoder compresses information in $\mathbf{X}_{\text{past}}$ to a latent space $\mathbf{z}_{AE}\in\mathbb R^{d_z}$. This compressed memory representation is passed as an input alongside present state variables $\mathbf{x}(t)$ to a NN, denoted by $\mathcal N$, that acts as a surrogate for unresolved processes and outputs the sub-grid scale contribution $B$. 
\begin{align}
    \mathbf{z}_{AE}&=\mathcal E_{M,d_z}(\mathbf{X}_{\text{past}}):\mathbb R^{M\times d_{x'}}\rightarrow \mathbb R^{d_z} \\
    B&=\mathcal N(\mathbf{z}_{AE}, \mathbf{x}(t)):\mathbb R^{d_z+d_x}\rightarrow \mathbb R 
\end{align}
Additionally, we train a decoder that converts latent space variables back to their original inputs $\mathbf{X}_{\text{past}}$. Even though the decoder is technically not necessary for parameterization, we found it important to enforce predictable latent dynamics. %

The proposed architecture for memory-informed ML parameterizations is illustrated in Figure~\ref{fig:architecture}. For one of our test cases, the encoder $\mathcal E_{M, d_z}$ and decoder $\mathcal D_{M, d_z}$ contain an optional 1d temporal convolution layer. Convolutional layers can improve model performance significantly by adding temporal connectivity to the input features \cite{beucler_distilling_2025}, but they also increase computational effort during training and inference. Whether or not the performance increase outweighs the computational overhead depends on the use case. The trainable parameters of the encoder, decoder and NN are learned simultaneously by optimizing the loss function
\begin{equation}
    \ell = \alpha \mathrm{MSE}(B, B^*) + (1-\alpha)\mathrm{MSE}(\mathbf{X}_{\text{past}},\mathbf{X}^*_{\text{past}})
\end{equation}
which consists of a prediction loss and a reconstruction loss that can be balanced with parameter $\alpha$. Target variables, gathered from the training data, are indicated by an asterisk. The reconstruction loss determines how well the decoder can reconstruct the initial input of the encoder $\mathbf{X}_{\text{past}}$ and the prediction loss quantifies how close the predicted subgrid-scale contribution $B$ is to the training target $B^*$. We did not evaluate sensitivity to $\alpha$ and set it to $\alpha=0.5$ in all our trained models. Our ML model is implemented using Pytorch \cite{ansel_pytorch_2024} and optimized using the Adam algorithm \cite{kingma_adam_2017}. Further information about the ML architecture is provided in \ref{app:hyper_opt}. 

\subsection{Symbolic Distillation of the Latent Space Dynamics} \label{sec:ED}
The goal of our equation discovery procedure is to replace the autoencoder by approximating the right-hand side (RHS) of an ODE
\begin{equation} \label{eq:ODE}
    \frac{\mathrm{d}\mathbf{z}}{\mathrm{d}t} = f(\mathbf{x}(t), \mathbf{z}(t))
\end{equation}
evolving the latent space variables $z$ in time. This step is a Markovianization of the memory contribution derived via the encoder, where the explicit dependence on past time steps is translated into the auto-regressive dynamics of $\mathbf{z}$. Further, we allow $f$ to depend on $\mathbf{x}(t)$ to stabilize the model, as $\mathbf{x}(t)$ is obtained from the simulation and does not propagate errors during offline training. To discover $f(\mathbf{x}(t), \mathbf{z}(t))$, we generate training data $\mathbf{z}_{AE}$ using the frozen encoder and compute their derivatives, $\mathbf{\dot z}_{AE}$, using central finite differences. These derivatives serve as the target variables for equation discovery.

During our experiments, we found that the chance of successfully finding $f$ depends strongly on the hyperparameters of the autoencoder, especially the number of past time steps to use $M$, the number of latent variables $d_z$, and the regularization strength $w$ (weight decay of the Adam optimizer). The autoencoder maps information from a high-dimensional ($M \times d_{x'}$) input space to a low-dimensional latent space. Evolving the latent space variables as a Markovian, autoregressive process with forcings might therefore be difficult, due to apparent stochasticity induced by information loss in the compression. Since symbolic equation discovery is highly sensitive to such noise and requires extensive tuning, we introduce a fast and robust proxy to assess whether a well-defined autoregressive mapping exists: we train a flexible neural network to approximate $f$ given the same inputs $\mathbf{x}(t), \mathbf{z}(t)$ as the equation discovery. By the universal approximation theorem \cite{hornik_multilayer_1989}, if a sufficiently expressive NN cannot learn this mapping, a symbolic regressor certainly will not. We quantify this latent predictability proxy using $R^2$ values of the NN predicting $\dot z_{AE,i}$ from $\mathbf{x}$ and $\mathbf{z}_{AE}$, denoted as $R^2_{\text{latent},i}(\dot z_{AE,i}, \dot z_{AE,i}^*)$. The $R^2_{\text{latent},i}$ is computed independently for each of the $d_z$ latent dimensions $i$. Because NN training on GPUs is fast compared to symbolic regression, $R^2_{\text{latent},i}$ provides an efficient, interpretable metric for autoencoder hyperparameter optimization.

Our tuning strategy balances two objectives: parameterization accuracy and latent space predictability. Parameterization performance (measured by $R^2(B, B^*)$ against the target subgrid-scale contribution) is strongly governed by $M$, so we first select the largest feasible $M$ that remains computationally tractable (see Table \ref{tab:hyper_opt_M}), as we observed a monotonic improvement with longer memory windows. With $M$ fixed, we then optimize $d_z$ and $w$ to maximize $R^2_{\text{latent,i}}$ while maintaining high parameterization performance. Since the latent space prediction skill is measured in $d_z$ dimensions, we monitor both the mean $\bar{R}^2_{\text{latent}}$ and the minimum $\min(R^2_{\text{latent, i}})$ over the entire latent space. The latter is important: because the latent variables evolve as a coupled system of ODEs, we anticipate that the worst-predicted dimension could act as a bottleneck, degrading the stability and accuracy of the entire symbolic model.

After finding a suitable set of hyperparameters, we attempt to find a symbolic description of equation \eqref{eq:ODE} describing the temporal evolution of the latent space variables from the autoencoder. 
We tested PySINDy \cite{silva_pysindy_2020, kaptanoglu_pysindy_2022} and Qlattice \cite{brolos_approach_2021} as complementary equation discovery backends. By evaluating both approaches we assess how the choice of equation discovery strategy influences the quality of the resulting prognostic models. We also briefly tested PySR \cite{cranmer_interpretable_2023} as an alternative to Qlattice, but did not obtain competitive results.

PySINDy transforms the equation discovery problem into a linear regression 
\begin{equation}
\frac{\mathrm{d}\mathbf{z}}{\mathrm{d}t} \approx \Theta(\mathbf{z},\mathbf{x})\mathbf{\Xi}
\end{equation}
by defining a library of generally nonlinear candidate functions $\Theta$ serving as potential building blocks of the equation. During the optimization procedure, coefficients $\mathbf{\Xi}$ are regularized to select a sparse combination of candidate functions. This results in a minimalistic set of equations that drive the ODE evolution, but the definition of a suitable candidate library $\Theta$ is essential for successful equation discovery. While PySINDy has emerged as a powerful tool in studying many systems in the field of nonlinear dynamics \cite{bakarji_discovering_2023, champion_data-driven_2019}, the candidate functions of a distilled NN are hard to guess because we miss physical theory guiding their functional form. As a consequence, symbolic distillation applications have moved to evolutionary algorithms \cite{tan_symtorch_2026}. Nonetheless we used PySINDy to test polynomial libraries up to fourth order. %

As an evolutionary alternative, the Qlattice framework combines different approaches from graph theory and evolutionary regression and is inspired by Richard Feynman's path integration formulation in quantum mechanics. It performed best in recent equation discovery benchmarks \cite{de_franca_srbench_2025} and is able to compose equations from fundamental mathematical operations instead of prescribing a candidate library. Qlattice samples a subset of possible equations from the QGraph, where equations are represented by unidirectional, acyclic graphs. Nodes in that graph, which represent variables, have weights and biases that are estimated using backpropagation. We will call those coefficients $\mathbf{\Xi}$ in analogy with PySINDy. After evaluating the performance of the different equations, with a loss function, Qlattice updates the underlying probability distribution of the QGraph, such that models with a lower complexity and higher predictive capabilities are favored in the next iteration. Over time, different paths in the QGraph emerge that approximate \eqref{eq:ODE}. Our Qlattice configuration uses the default Qlattice setup and trains for 1000 epochs, while reducing the size of the training data set to $5\cdot10^6$ samples. This is commonly done in equation discovery, as the number of trainable parameters is small compared to ML problems, and memory consumption is critical when evaluating many equations in parallel.

The predicted latent space variables $\mathbf{z}_{ED}$ evolved using the discovered equation are not perfectly equivalent to the ones predicted by the autoencoder $\mathbf{z}_{AE}$. To give the parameterization $\mathcal N$ a chance to adapt to the new set of latent space variables, we had to retrain $\mathcal N$ with the new inputs $\mathbf{z}_{ED}$. To further improve model performance, we then freeze the structure of the discovered equation, but pass its constants $\mathbf{\Xi}$ as trainable parameters to the optimization process of the parameterization. Since we need to evolve the ODE during training, we implemented an Euler step integration scheme as part of the forward call of our model. In our experiments, we applied a roll-out training technique, where the ODE is integrated for a short time period before being re-initialized with latent space variables from the autoencoder. The roll-out times are increased in subsequent training epochs until the ODE is integrated over the whole temporal domain. This approach delivered better results than immediately integrating the ODE over the whole training set. The graph structure of Qlattice, with weights and biases attached to variable nodes, results in more trainable parameters, giving the equation more flexibility compared to frameworks like PySINDy where coefficients within the candidate functions are not possible, and PySR, which we found to create few constants during equation discovery. We provide additional information on how to obtain stable roll-out-training in the supplementary material.

\subsection{Lorenz-96 Model} \label{sec:methods_L96}
To assess the capabilities of our framework, we first implement it with the Lorenz-96 (L96) model \cite{lorenz_predictability_1995}. This model has been used in several studies as a proof-of-concept to investigate different research paths concerning parameterizations for ESMs \cite{gagne_ii_machine_2020, rasp_coupled_2020}. The reason why L96 is used frequently in this domain is that it replicates the structural challenges of ESMs, while implementation is easy and simulations can be performed quickly. L96 shows chaotic dynamics of self-advecting, multi-scale momentum variables and the model time unit (MTU) can be directly linked to atmospheric time scales of $1\,\text{MTU}\approx5\,$d by comparison of error doubling times. We will focus on the two-level implementation, which consists of slow coarse-scale variables $X_k$, that are bi-directionally coupled to the fast evolving variables $Y_{j,k}$ and can be described by a set of coupled ODEs.
\begin{align} \label{eq:L96_memory}
    \frac{\mathrm{d}X_k}{\mathrm{d}t} &= 
    -X_{k-1} (X_{k-2} - X_{k+1})
    - {X_k}
    +F
    \underbrace{-\frac{hc}{b}\sum_{j=1}^{J}Y_{j,k}}_{\text{Coupling} \equiv B} \\
    \frac{1}{c} \frac{\mathrm{d}Y_{j,k}}{\mathrm{d}t} &= 
    -b Y_{j+1,k}(Y_{j+2,k} - Y_{j-1,k})
    -Y_{j,k}+
    \frac{h}{J} X_k
\end{align}
We set the forcing $F=20$, coupling strength $h=1$, relative evolution speed $c=10$, fast advection magnitude $b=10$, number of slow variables $K=8$ and number of fast variables $J=32$. This parameter configuration seperates the attractor into two regimes \cite{christensen_simulating_2015}, which we want to use for our evaluation. 
To simulate L96 in time, we use a four step Runge-Kutta (RK4) integration scheme with step size $\Delta t=0.001\,$MTU. Our software implementation was adapted from \citeA{rasp_coupled_2020}.

To create training data, we simulate L96 for 10,000 MTUs and split the data along the time dimension into two equal-sized train and test datasets. The autoencoder uses past time steps of the local $X_k$ variable to create latent space variables, which are used together with the present time $X_k$ variable to predict the coupling term $B$ with the parameterization.

Finally, we couple our parameterizations to L96. %
For initialization, the standard L96 model is integrated using the full equations for one MTU. From this time onward, the $Y_{j,k}$ are turned off and the sub-grid scale contributions $B$ are parametrized. In the case of the ODE-based parameterization, another MTU is simulated using the autoencoder, and $\mathbf{z}_{AE}(t_0=2\,\text{MTU})$ is used as the initial condition for the ODE. Each L96 version (original or with one of the parameterizations) is integrated for 50,000\,MTU to generate evaluation data. Additionally, an ensemble of 500 short simulations with different initial conditions was generated for each version to assess its "weather" prediction capability. To ensure independence, the initial conditions for this ensemble were drawn from the full simulation with a time interval of 10 MTU. 

\subsection{Precipitation parameterization}
As our second experiment we predict coarse-grained precipitation from a high-resolution ICON simulation produced by the ICON-NWP setup as part of the DYAMOND-winter model inter-comparison project \cite{stevens_dyamond_2019}. ICON solves a set of non-hydrostatic primitive equations on an icosahedral grid. The global grid, used for the simulation, has an approximate horizontal resolution of 2.5$\,$km and 90 vertical layers. The simulation covers a 30$\,$d period in February, after 10$\,$d of spin-up, with an output frequency of 15 minutes for 2d variables. To create a parameterization setup, we coarse-grain the high-resolution data to a horizontal resolution of 80$\,$km, comparable to state-of-the-art ESM resolutions for climate prediction. We adopted the coarse-graining pipeline from \cite{grundner_data-driven_2024}, utilizing the CDO library \cite{schulzweida_cdo_2023}.\\ %

As inputs to our precipitation parameterization, we use the following 2d fields
\begin{equation}
    \textbf{X}_{\text{I}} = \{q_{\text{2m}}, PW, T_{\text{2m}}, T_{\text{sfc}}, H, LE, l\}
\end{equation}
combining the two meter specific humidity $q_{\text{2m}}$, column-integrated precipitable water $PW$, two meter air temperature $T_{\text{2m}}$, surface temperature $T_{\text{sfc}}$, sensible heat flux at the surface $H$, latent heat flux at the surface $LE$, and a binary land-sea-mask $l$. During pre-processing, we standardize our input data. We used this subset of parameters because they are available with a high temporal output frequency of 15 minutes and because similar subsets have been used in the past for demonstrative purposes of predicting precipitation \cite{beucler_distilling_2025, shamekh_implicit_2023}. We separate the data to use the first $\approx23\,$d for training, the following $\approx4\,$d for validation, and the final $\approx4\,$d for our evaluation. To create the input variables $\mathbf{X_{\text{past}}}$ for the autoencoder, we included $M$ past time steps for all variables in $\textbf{X}_{\text{I}}$ except $l$. %

\subsection{Baseline Models}\label{sec:baseline_models}
To compare the performance of our memory-informed ML parameterizations, we additionally train a local and a non-local (only for L96) baseline neural network that takes only the present time steps $\mathbf{x}(t)$ as an input. Because $\mathbf{X}_{\text{past}}$ is essentially a time-delayed embedding with embedding dimension $M$ and embedding delay $\Delta t$, we want to ensure that the memory models learn more than just non-local information. This is important as we know from Takens Theorem \cite{takens_detecting_1981}, that any sufficiently parametrized time-delayed embedding can be mapped to the full attractor via a diffeomorphic transformation. Autoencoders have been used in combination with time-delayed embeddings to achieve that \cite{bakarji_discovering_2023}, showing that the full set of governing equations, from another Lorenz system (L63), can be retrieved from a single observable using equation discovery. %
To quantify how much improvement comes from non-local information, we compare our L96 memory models to a local NN and a non-local NN that takes all $\{X_{k=1}(t), ..., X_{k=8}(t)\}$ as input.

For the km-scale precipitation parameterization, because adding non-local information beyond nearest neighbors is much more difficult to implement in operational ESMs than carrying prognostic memory variables, we refrain from comparing our memory parameterizations to a non-local baseline. Nevertheless, offline parameterizations for convection have demonstrated that adding inputs from nearest neighbors improves the parameterization in cases of mesoscale convective organization beyond a single grid cell \cite{wang_non-local_2022}, further highlighting the connection between non-locality in space and time. We also do not evaluate our parameterization online, as coupling the equation to the dynamical core of an ESM is beyond the demonstrative purpose of this work.  

\section{Lorenz-96 Results} \label{sec:L96}
\begin{figure}[t]
    \centering
    \includegraphics[width=1.0\linewidth]{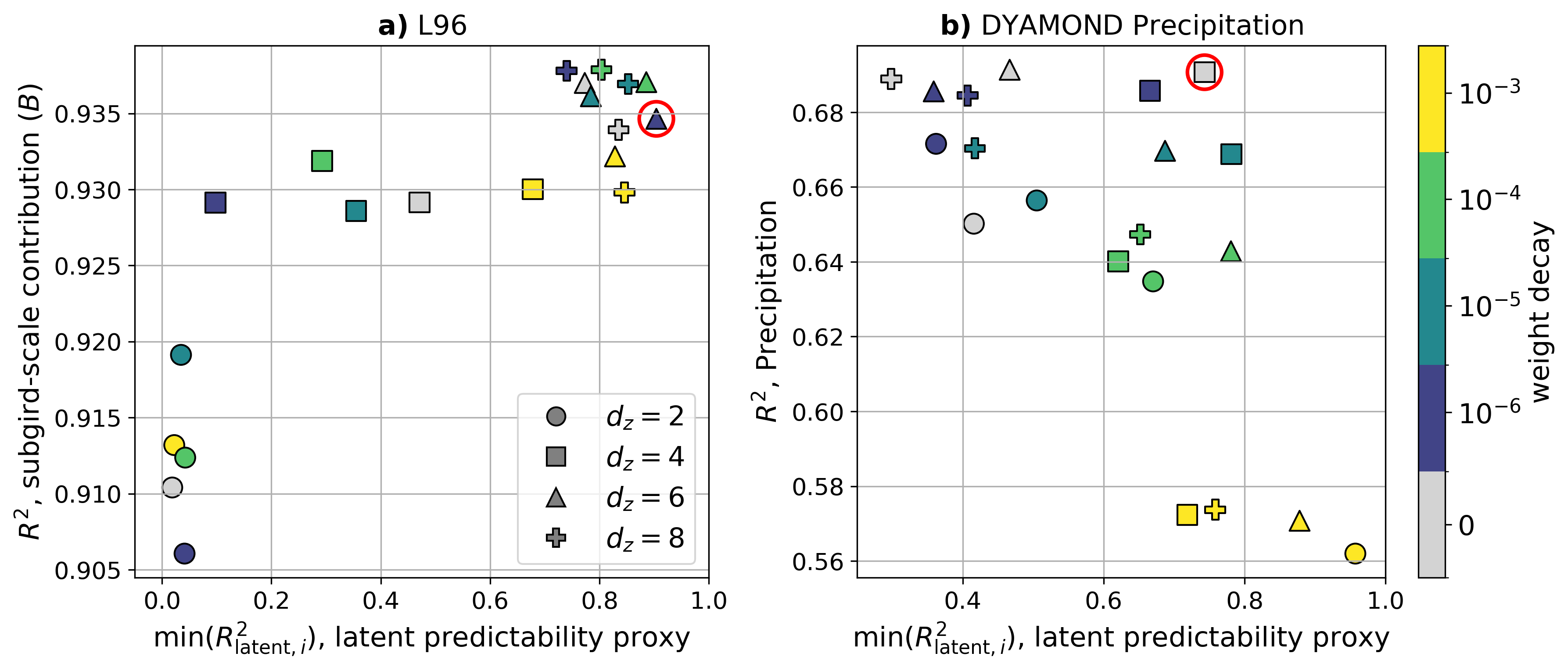}
    \caption{Hyperparameter optimization to balance prediction accuracy of the parameterization (measured with $R^2$) and the latent predictability proxy derived via a Markovian NN (measured with $\min(R_{\text{latent},i}^2)$ of the worst latent space dimension). Panel \textbf{a)} shows L96 and panel \textbf{b)} shows  the precipitation experiment. Selected models are marked with a red circle.}
    \label{fig:hyper_opt}
\end{figure}
We performed hyperparameter optimization as described in section \ref{sec:ED}. We employ two skill measures: the overall parameterization accuracy as measured by $R^2$, and the latent predictability proxy, measured by $R^2_{\text{latent},i}$.  Figure \ref{fig:hyper_opt} shows the impact of the autoencoder's latent space dimension $d_z$ and regularization strength (weight decay $w$). 

The latent space dimension has a strong impact on both parameterization performance and the latent predictability proxy which suffers from over compressing the latent space $d_z<6$. In addition, the different $w$ model realizations show significant variance in $R^2$ for $d_z=2$ and $\min(R_{\text{latent},i}^2)$ for $d_z=4$. 

We chose the model that maximized $\min(R_{\text{latent},i}^2)$ and still shows reasonable parametrization accuracy. The selected model, indicated by the red circle, has hyperparameters $d_z=6,\, M=1000,\, w=10^{-6}$. Our equation discovery setup, utilizing PySINDy, finds that a linear ODE emulating the selected autoencoder performs sufficiently well, achieving an $R^2=0.86$ when predicting $\dot{\mathbf{z}}_{AE}$ compared to the latent predictability proxy of $\bar R_{\text{latent}}^2=0.935$. Given the strong performance of the linear ODE and its simple structure, we do not perform roll-out training or attempt to further improve emulation performance with Qlattice. Instead, we simply retrain the NN parametrization using the latent space variables $\dot{\mathbf{z}}_{ED}$ generated by the linear ODE. In the following, we discuss the performance of the memory-informed parameterizations (where the autoencoder is replaced by the linear ODE) using online evaluation metrics, and subsequently compare them to the baseline models before interpreting the ODE.

\subsection{Climate Evaluation}
\begin{figure}[tbp]
    \centering
    \includegraphics[width=1.0\linewidth]{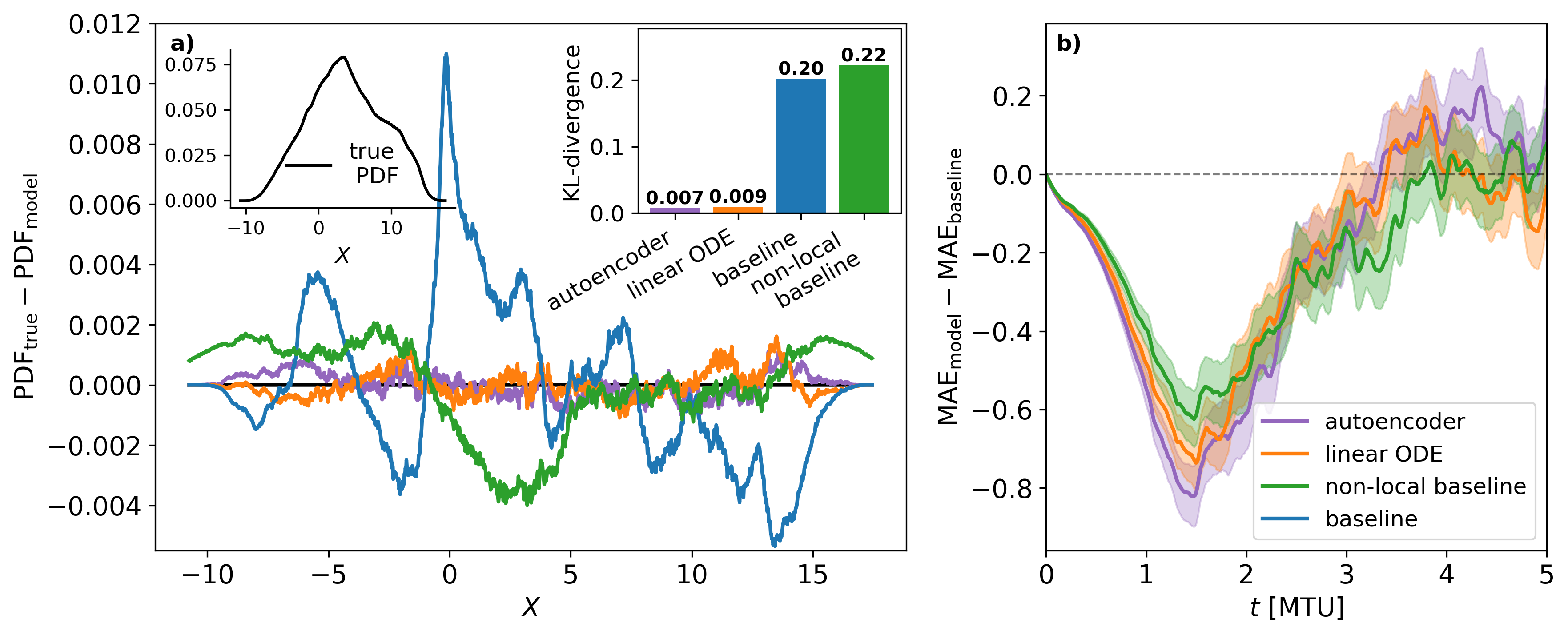}
    \caption{Prognostic memory variables improve distributions of state variables, leading to an accurate L96-climate, with (a) showing the difference between the full simulation PDF (left inset) and the parameterized runs. Climate scores are computed as Kullback-Leibler divergence values between true and parameterized simulation (right inset). Weather prediction capabilities show that prognostic memory is more effective than non-local information, reducing MAE when compared to the local baseline (b). Shaded areas show standard errors.}
    \label{fig:L96_climate_weather}
\end{figure}

For climate applications, the parameterization needs to push the coarse-scale model towards the accurate distribution of each coarse-scale variable. To analyze this capability for our different parameterizations, we compute histograms of the $X_k$ states for a 50,000\,MTU-long online simulations. The bin size for our ground truth L96 simulation is estimated with the Freedman–Diaconis rule \cite{freedman_histogram_1981}, resulting in 1404 bins. To quantify the difference between our parametrized simulation $Q_{\text{model}}$ and the ground truth simulation $P$, we use the well known KL-divergence.
\begin{equation}
    D_{\mathrm{KL}}(Q_{\text{model}}) = \sum_{x \in X}P(x)\log\frac{P(x)}{Q_{\text{model}}(x)}
\end{equation}
The evaluation results are shown in the inset of Figure \ref{fig:L96_climate_weather}.

The best climate is produced by the memory-informed parameterization with latent space variables generated by the autoencoder with $D_{\mathrm{KL}}=0.007$, followed by the distilled model recovering most of the climate score $D_{\mathrm{KL}}(Q_{\text{linear ODE}})=0.009$. In comparison, both baseline models lead to significantly worse predictions, with similar KL-divergence values $D_{\mathrm{KL}}\approx0.2$. Further, the memory-informed parameterizations improve climate scores compared to both the Generative Adversarial Network and the Recurrent-Neural-Network (RNN) with gated recurrent units, as trained in \citeA{parthipan_using_2023}. The latter is especially interesting as the RNN emerged as the best model architecture, and it suggests that the prognostic memory variables are able to leverage information from the past time steps that the RNN misses. This results in a much lower KL-divergence ($D_{\mathrm{KL}}(Q_{\mathrm{RNN}})=0.04$), even though the gated recurrent units in the RNN are designed to carry long-term memory information. Because KL-divergence depends on the binning strategy, we also computed KL-divergence for 827 bins as in \citeA{parthipan_using_2023}, which further increases model discrepancy as our memory-informed parameterizations improve their climate scores.%

For comparison, \citeA{brolly_stochastic_2025} reports that a non-local mixture-density network (MDN) improves upon its local counterpart in climate evaluations. In contrast, adding non-local information to our deterministic parameterizations does not improve climate scores and instead leads to an overprediction of the distribution tails, slightly increasing the KL-divergence. However, both studies find that introducing memory improves the performance of local parameterizations.

\subsection{Weather Evaluation}
To analyze the weather prediction capabilities of the different parameterizations, we compute the mean absolute error $\text{MAE}$ between 500 7\,MTU-long parametrized runs and the full system simulation (switching in the parameterizations after two MTU as noted in Section~\ref{sec:methods_L96}). 
\begin{equation}
    \text{MAE}_{\text{model}}(t) = \frac1K \sum_{k=1}^K \vert X^{\text{model}}_k(t) - X^{\text{truth}}_k(t) \vert
\end{equation}

Results are visualized in Figure \ref{fig:L96_climate_weather} comparing $\text{MAE}_{\text{model}}(t)$ against the local baseline. All models demonstrate better forecast skill compared to the baseline NN, with maximum benefit at 1-2 MTU, which corresponds to 5-10 atmospheric days. Benefits vanish at a lead time of around 3-4 MTU or 15-20 atmospheric days. Even though the non-local baseline could not improve upon the local baseline in recreating the correct state distributions, adding non-local information seems to inform the parameterization about spatial patterns, which improve short-term predictions, but to a lesser extent than the memory-informed models. Similar improvements of memory-informed parameterizations, in the context of short-term forecasts, were shown by \citeA{bhouri_memory-based_2023}. They additionally showed that memory-informed parameterizations generalize better to unseen forcing regimes, by varying $F$ in equation \eqref{eq:L96_memory}.

The importance of non-locality for weather evaluation of L96 was also shown for stochastic MDNs in \citeA{brolly_stochastic_2025}. Further, their local parameterization improves with respect to weather prediction by adding memory and performs similarly to their non-local parameterization without memory, consistent with our results.

\subsection{Regime Analysis}
\begin{figure}[tb]
    \centering
    \includegraphics[width=1.0\linewidth]{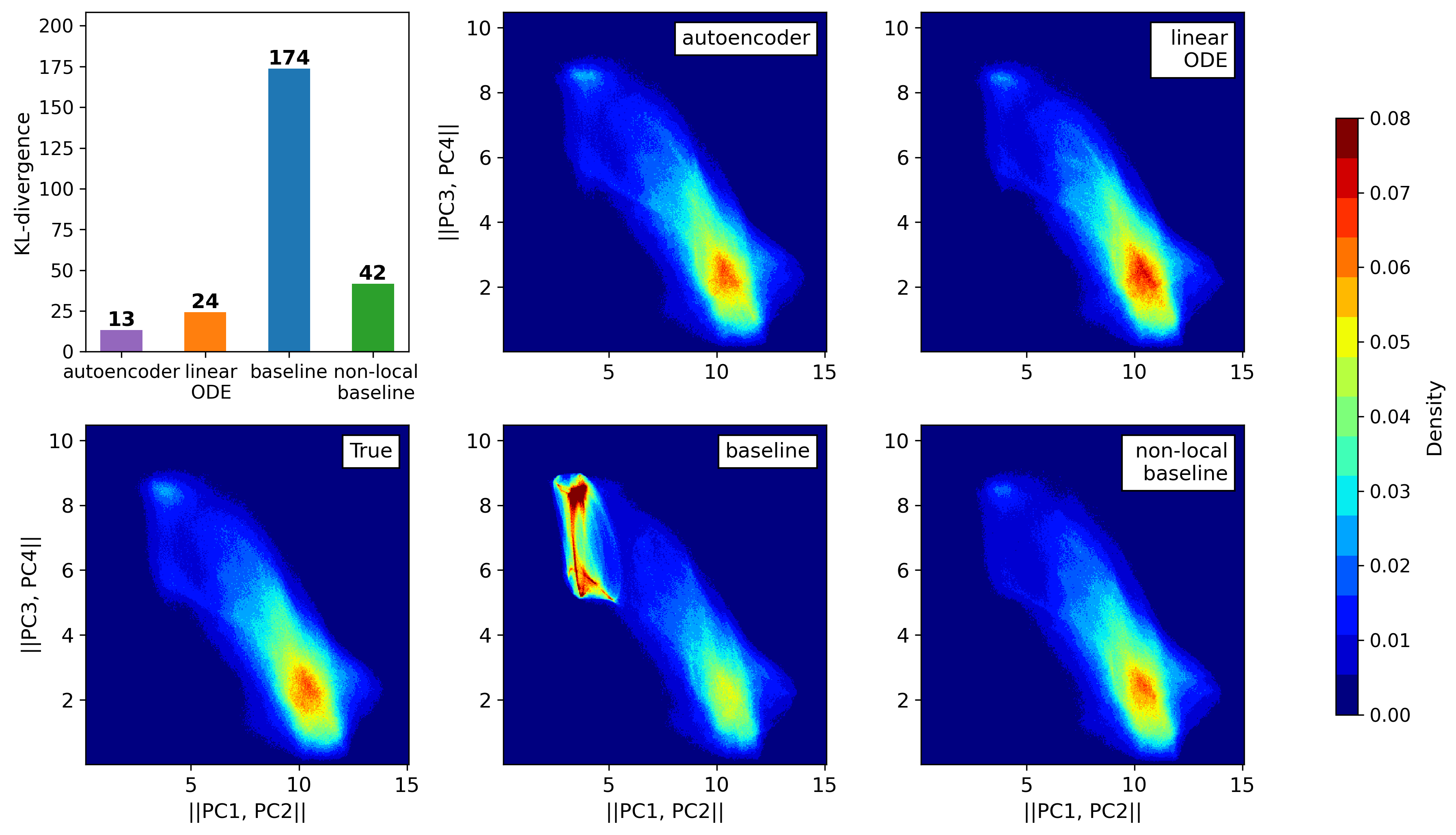}
    \caption{Prognostic and non-local parameterizations capture the overall two-regime structure of the L96 dynamics, improving upon the local in time and space baseline. The difference to the true distribution is quantified by the bar plot in the top left showing Kullback-Leibler divergence values.}
    \label{fig:L96_regimes}
\end{figure}
The L96-configuration used in our study leads to two regimes that dominate the dynamics of L96 in the form of wave patterns with wavenumber 1 (type-1) and 2 (type-2) \cite{christensen_simulating_2015}. %
To identify regimes, we follow the approach described in \citeA{parthipan_using_2023}, and decompose $X(t)$ into four principal components (PCs). We combine $PC_1, PC_2$ and $PC_3, PC_4$ as they correspond to the same phase-shifted waves. 
\begin{align}
    \vert\vert PC_1,PC_2\vert\vert &= \sqrt{PC_1^2+PC_2^2}\\
    \vert\vert PC_3,PC_4\vert\vert &= \sqrt{PC_3^2+PC_4^2}
\end{align}
We show 2d-histograms of the combined PCs and KL-divergence between the ground truth and parametrized simulation in Figure \ref{fig:L96_regimes} with 350 bins to match the procedure in \citeA{parthipan_using_2023}. All parameterizations, except the local baseline, are able to capture the two-regime structure of our L96 configuration. The autoencoder shows the lowest KL-divergence with $D_{\mathrm{KL}}=13$, followed by the linear ODE with $D_{\mathrm{KL}}=24$. Both our memory models outperform the RNN from \citeA{parthipan_using_2023}, whose KL-divergence was $D_{\mathrm{KL}}=32$, and both baselines. Visually, the linear ODE model slightly underestimates the less frequent type-1 regime in favor of the dominant type-2 regime. %
However, the linear ODE shows similar characteristics to the autoencoder, showcasing that a linear ODE with memory of past states can reproduce the chaotic regimes of L96 beyond non-local information and state-of-the-art RNNs.

The non-local baseline ($D_{\mathrm{KL}}=42$) clearly improves upon the local baseline ($D_{\mathrm{KL}}=174$). The structurally distinct distribution of the local baseline visually matches that obtained from deterministic third-order polynomial parameterizations with lightly perturbed coefficients \cite{christensen_simulating_2015}. Since the third-order polynomial already provides a good approximation, it is plausible that the weak higher-order structures learned by the NN act as small corrections to the cubic representation. This may explain why the observed regimes resemble the perturbed parameterizations reported in \citeA{christensen_simulating_2015} rather than the corresponding best-fit solution.

It seems intuitive that non-local information would enable a parameterization to distinguish among large-scale wave configurations in the system. In that case, the parameterization can condition its estimate of the unresolved sub-grid scale contributions on the current synoptic state, leading to a more accurate representation of the regime distribution. This interpretation is consistent with the improved regime statistics obtained for the non-local baseline. Since these synoptic states are persistent features of the large-scale dynamics, their improved representation may also contribute to the enhanced short-term forecast skill shown for non-local parameterizations. However, the present results suggest that these benefits do not necessarily translate into improved climate statistics, indicating that accurate regime representation alone may be insufficient for reproducing the long-term climatology. An exception might be the correct estimation of extreme events.

\subsection{Symbolic Prognostic Memory Variables for L96} \label{sec:ODE_interpretation_L96}
The linear ODE takes the general form  
\begin{equation} \label{eq:linear_ODE}
    \frac{\mathrm{d}z_l}{\mathrm{d}t} = 
    \xi_{l, 0} + \xi_{l, 1}z_{1} + ... + \xi_{l, d_z}z_{d_z} + \xi_{l,d_z+1}x_1(t) + ... + \xi_{l,d_z+d_x}x_{d_x}(t)
\end{equation}
with $d_z$ self-propagating equations and an inhomogeneous part consisting of the $d_x$ forcing variables and a constant.

To derive a semi-analytic solution for our prognostic memory variables, we write the full system of ODEs as
\begin{equation}\label{eq:linear_ODE_matrix}
    \frac{\mathrm{d}\mathbf{z}}{\mathrm{d}t}
    = \mathbf{\Xi} \begin{pmatrix}
1\\
\mathbf{z}\\
\mathbf{x}
\end{pmatrix} =  \mathbf{\Xi}_b + \mathbf{\Xi}_z \mathbf{z} + \mathbf{\Xi}_x \mathbf{x} 
\end{equation}
with
\begin{equation}
\mathbf{\Xi}=
\begin{pNiceMatrix}
\xi_{1,0} & \xi_{1,1} & \dots & \xi_{1,d_z} & \xi_{1,d_z+1} & \dots & \xi_{1,d_z+d_x} \\
\vdots & \vdots & \vdots & \vdots & \vdots & \vdots & \vdots \\
\xi_{d_z,0} & \xi_{d_z,1} & \dots & \xi_{d_z,d_z} & \xi_{d_z,d_z+1} & \dots & \xi_{d_z,d_z+d_x}
\CodeAfter
\UnderBrace[yshift=4pt]{1-1}{3-1}{\mathbf{\Xi}_b}
\UnderBrace[yshift=4pt]{1-2}{3-4}{\mathbf{\Xi}_z}
\UnderBrace[yshift=4pt]{1-5}{3-7}{\mathbf{\Xi}_x}
\end{pNiceMatrix}.
\end{equation}

\vspace{8pt}
The set of equations in \eqref{eq:linear_ODE_matrix} is known as a linear-time-invariant system with forcing and has the general solution
\begin{equation}\label{eq:linear_dZdt_solution}
    \mathbf{z}(t)=e^{\mathbf{\Xi_z} \left(t-t_0\right)}\mathbf{z}(t_0) + \int_{t_{0}}^t e^{\mathbf{\Xi_z}(t-s)}(\mathbf{\Xi_x}\ \mathbf{x}(s) + \mathbf{\Xi_b})ds.
\end{equation}
This solution is guaranteed to have a stable trajectory if $\mathrm{Re}(\lambda_i)<0$ for all eigenvalues $\lambda_i$ of $\mathbf{\Xi_z}$. From \eqref{eq:linear_dZdt_solution} we can see that our prognostic memory variables are independent of their initial conditions for large $t$, which means they could be initiated without initial conditions from the autoencoder. In an ESM setup, this might simplify the implementation of prognostic memory variables, as  tracking past time steps and inferring their initial conditions with the autoencoder would otherwise require additional steps. The integral in \eqref{eq:linear_dZdt_solution} integrates pulses formed by linear combinations of the forcing variables and evolved by the latent dynamics. Parameters learned in the latent space only influence the evolution of the system by specifying the behavior of these pulse evolutions.\\
Due to the simple structure of linear ODEs, we can interpret the solutions in terms of eigenmodes.
\begin{figure}[tb]
    \centering
    \includegraphics[width=\linewidth]{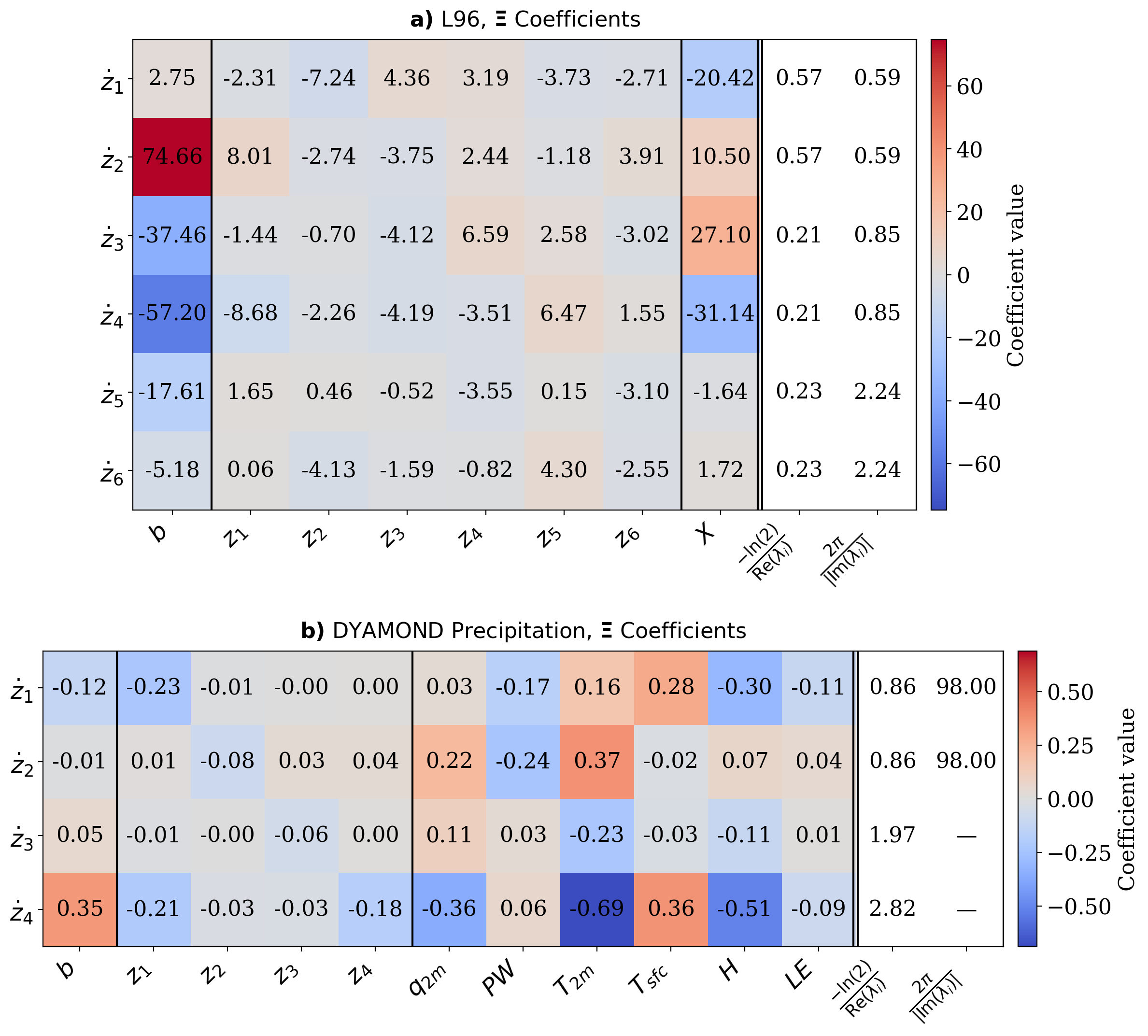}
    \caption{The magnitude of coefficients $\mathbf{\Xi}$ forming the linear ODE set are shown for the different sub-matrices defining the constant part $\mathbf{\Xi}_b$, the latent dynamics $\mathbf{\Xi}_z$ and the forcing composition $\mathbf{\Xi}_x$ (separated by solid lines). The prognostic variables learn to delay coarse-scale inputs to best parameterize the subgrid-scale contribution, following eigenmodes with half-life time and oscillation period separated by double-lines and given in MTU for L96 (a) and hours for the precipitation experiment (b).}
    \label{fig:Xi_L96_DYAMOND}
\end{figure}
The coefficients $\mathbf{\Xi}$ and eigenvalues of $\mathbf{\Xi}_z$ are shown in Figure \ref{fig:Xi_L96_DYAMOND}. To analyze the eigenvalues as more meaningful quantities, Figure \ref{fig:Xi_L96_DYAMOND} also displays the half-life $t^{(i)}_{1/2}$ of the exponential decay and the oscillation period $T^{(i)}$, 
\begin{align}
    t^{(i)}_{1/2}&=-\frac{\ln2}{\mathrm{Re}(\lambda_i)} \\
    T^{(i)}&=\frac{2\pi}{|\mathrm{Im}(\lambda_i)|}
\end{align}
in case of complex eigenvalues.

For the L96 application, we find that $\mathrm{Re}(\lambda_i)<0$, guaranteeing stable trajectories, and that all six latent space variables come as complex conjugate pairs (Figure \ref{fig:Xi_L96_DYAMOND}a). Paired variables follow a decaying rotation dynamic in the eigenbasis, where the real part of the eigenvalue gives the decay rate and the imaginary part gives the rotation frequency. Those dynamics describe how signals from the forcing travel around the L96-system and periodically impact the latent space variables. This behavior could reflect the self-advecting properties of L96 as well as the periodic boundary conditions. %
As seen in Figure \ref{fig:Xi_L96_DYAMOND}a, the slowest decaying mode ($t^{(1,2)}_{1/2}=0.57$\,MTU) has the fastest rotation ($T^{(1,2)}=0.59$\,MTU), showing that $\approx50\%$ of a forcing pulse's magnitude has decayed after a full rotation. In contrast, the mode corresponding to the eigenvalues $\lambda_{5,6}$ has negligible rotational dynamics as the decay $t^{(5,6)}_{1/2}=0.23$\,MTU is much faster than the rotation $T^{(5,6)}=2.24$\,MTU.

\section{Parametrizing Precipitation from Coarse-Grained Storm-Resolving Simulations} \label{sec:DYAMOND_precip}

\begin{figure}[tb]
\begin{minipage}[t]{0.45\linewidth}
    \centering
    \includegraphics[width=\linewidth]{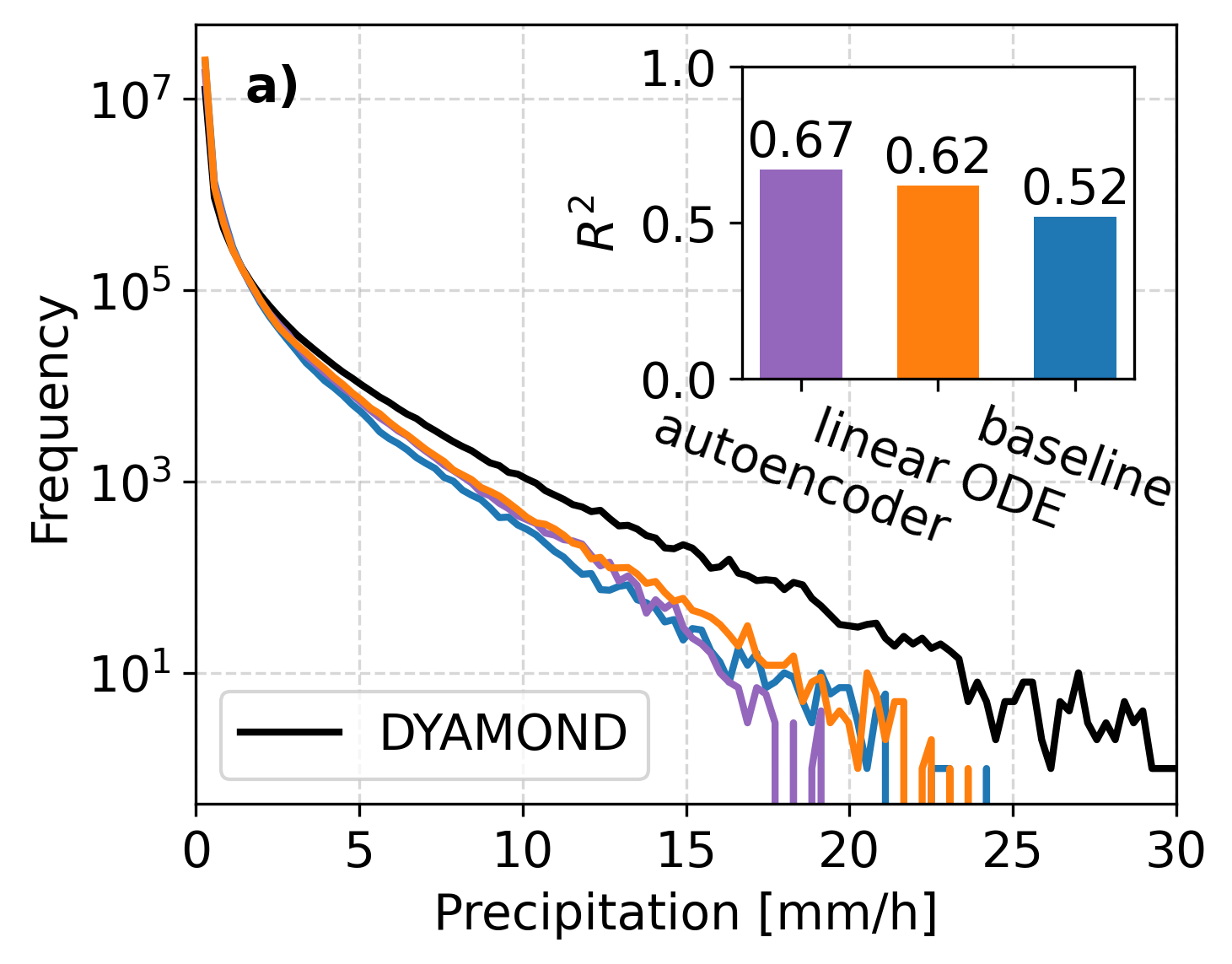}
\end{minipage}
\begin{minipage}[t]{0.55\linewidth}
    \centering
    \includegraphics[width=\linewidth]{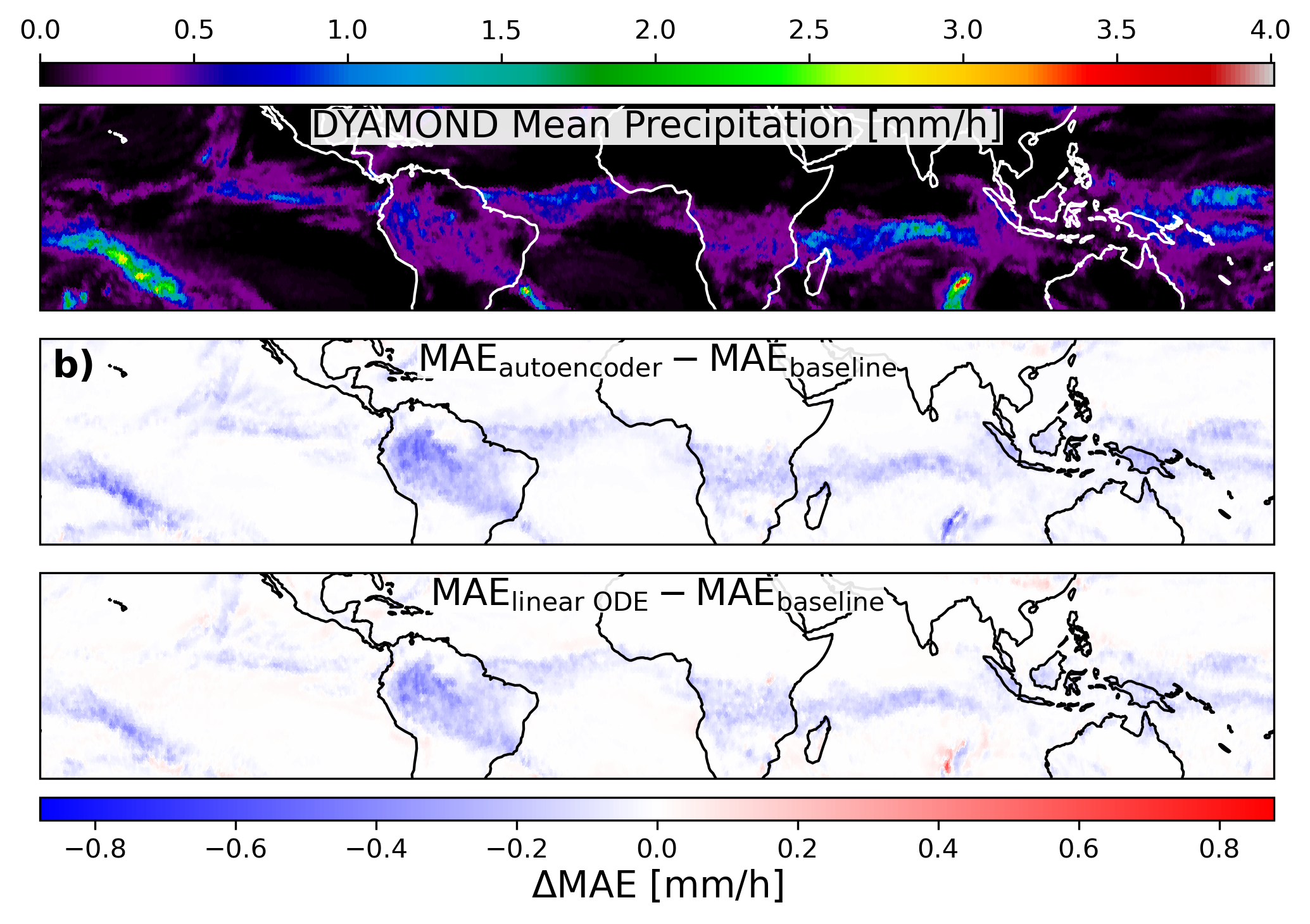}
\end{minipage}
\caption{A linear ODE recovers a significant amount of added value discovered by the memory-autoencoder over the diagnostic baseline parameterization (inset a). While the overall precipitation distributions appear similar (a), strong precipitation events in the tropics are improved (b).}
\label{fig:DYAMOND_precip}
\end{figure}

As expected, precipitation is much more challenging to parameterize than L96. We again provide sensitivity of the parameterization accuracy against the latent predictability proxy, for all tested hyperparameters in Figure \ref{fig:hyper_opt}b. In contrast to L96, we observe a clear trend how weight decay impacts our framework, where stronger regularization tends to simplify the latent space making it more predictable at the cost of parameterization performance. We chose the autoencoder with hyperparameters $d_z=4,M=20,w=0$ as our best model. The latent predictability proxy of this model gave $\bar R^2_{\text{latent}}=0.82$, while the worst predicted latent space variable has $\min(R^2_{\text{latent},i})=0.74$, potentially limiting the accuracy of all coupled latent space variables. Our best performing ODE, generated by Qlattice, could replicate the latent space variables of the autoencoder $\dot{\mathbf{z}}_{AE}$ with $R^2=0.65$ compared to a linear ODE, which resulted in $R^2=0.50$. Indicating complicated latent dynamics, which the NN computing the predictability proxy replicates, but are challenging for equation discovery approaches. Due to the large discrepancy between the latent space variables generated by the autoencoder and the ODEs, we performed roll-out training as described in \ref{sec:ED} for the nonlinear ODE generated by Qlattice and the linear ODE. Despite demonstrating lower skill when emulating the autoencoder, the linear ODE showed better results than the nonlinear ODE after roll-out training. We will therefore focus our evaluation on the prognostic memory variables generated by the linear ODE and provide additional evaluation for the nonlinear ODE in the supplementary material. As for the L96 we first evaluate the memory-informed parameterizations (autoencoder and linear ODE) against the baseline and then provide an interpretation of the ODE.

\subsection{Performance Metrics and Distribution Analysis}

For the evaluation of our different parameterizations, we use four days of unseen test data. Fully informing the parameterization about past observables increases $R^2$ from $0.52$ for the baseline parameterization to $0.67$ (inset of Figure~\ref{fig:DYAMOND_precip}a). Comparable improvements have been shown in \citeA{beucler_distilling_2025} with a similar experimental setup. The linear approximation of the latent dynamics ($R^2=0.62$) recovers a significant part of the added value discovered by the autoencoder.

We compute the precipitation distribution using the test period of four days, which is sufficient to capture the overall structure of the precipitation distribution but too short to robustly estimate the tails, as rare, high-intensity events are unlikely to be adequately sampled. As a potential consequence all models underestimate strong rain rates. This issue might further be impacted by the limited set of input features, as predicting intense precipitation from just 2d-fields is challenging. The low to strong precipitation regimes ($P<15\,\text{mm}\text{h}^{-1}$) show small improvements with the memory-informed parameterizations, less apparent due the logarithmic scale of the y-axis. %

\subsection{Spatial and Temporal patterns of Memory-Informed Parameterizations}
We next evaluate the spatial pattern of the model climatologies. Figure~\ref{fig:DYAMOND_precip}b explains that model improvements, shown by $R^2$ values, come from better predictions of strong precipitation events in the tropical belt, where there is little difference between the autoencoder model and the linear distillation. Combined with our interpretation of the linear coefficients in \ref{sec:ODE_interpretation_DYAMOND}, our results suggest that the linear ODE improves upon the baseline by learning physical processes, like convective organization, from past atmospheric states, critical to improve tropical precipitation estimates in ESMs. Similar conclusions were drawn by \citeA{shamekh_implicit_2023} who explicitly targeted spatial organization using an autoencoder trained on the high-resolution dataset.

In the supplementary material, we show that precipitation predictions over the entire dataset ($\approx30\,$d), initialized without the autoencoder as $\mathbf{z}(t_0)=\mathbf{0}$, show no sign of drifting towards a regime where information derived from the linear latent dynamics becomes less informative (see Figure S4 in the supplementary material). Even though those results were expected from the semi-analytical solution of the linear ODE set \eqref{eq:linear_dZdt_solution}, we see no temporal drift of the MAE for the best nonlinear ODE as well. This indicates that initializing the latent evolution process without access to the autoencoder might be possible even in more complicated setups.

\begin{figure}
    \centering
    \begin{minipage}[t]{0.49\linewidth}
        \centering
        \includegraphics[width=\linewidth]{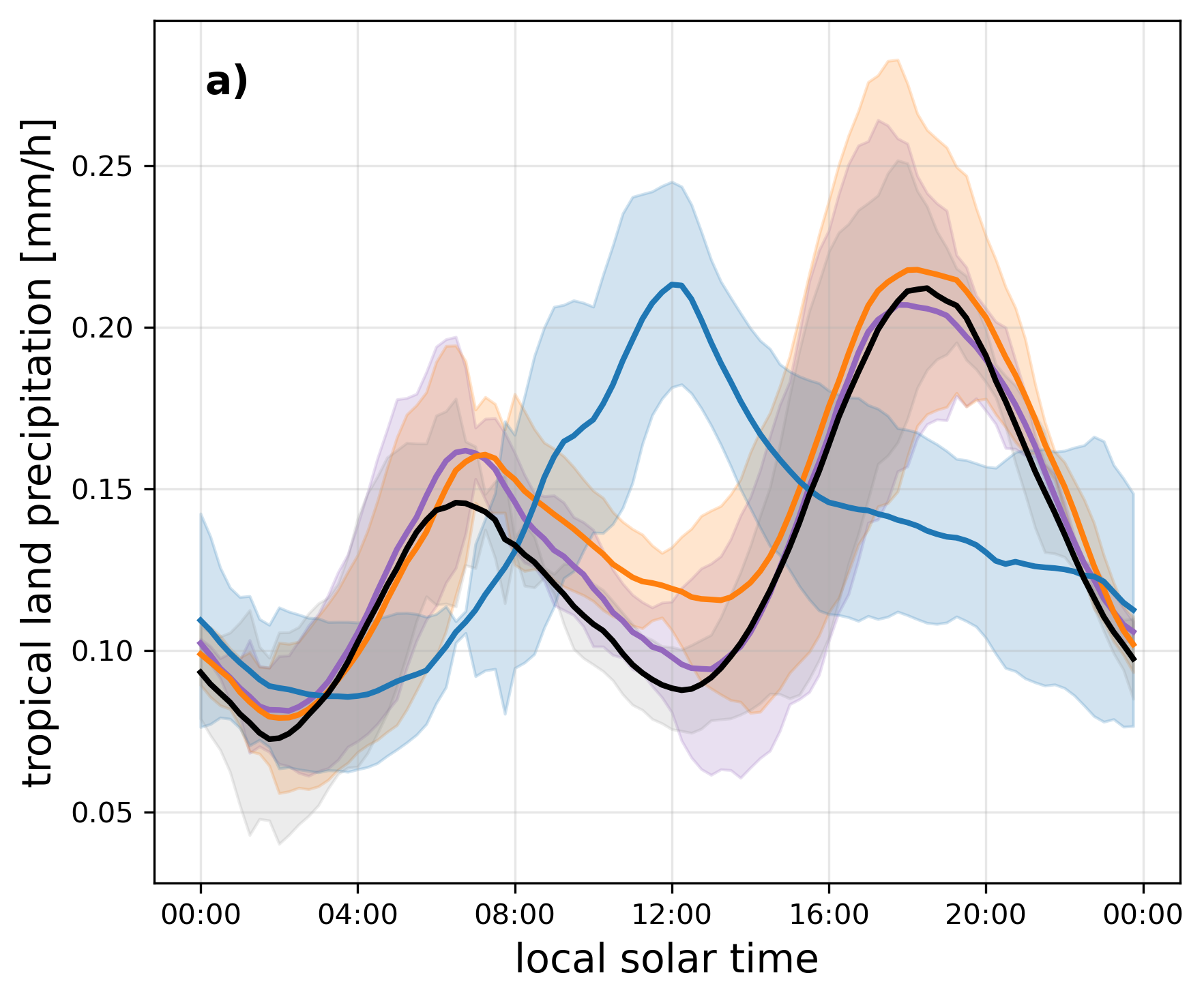}
    \end{minipage}
    \begin{minipage}[t]{0.49\linewidth}
        \centering
        \includegraphics[width=\linewidth]{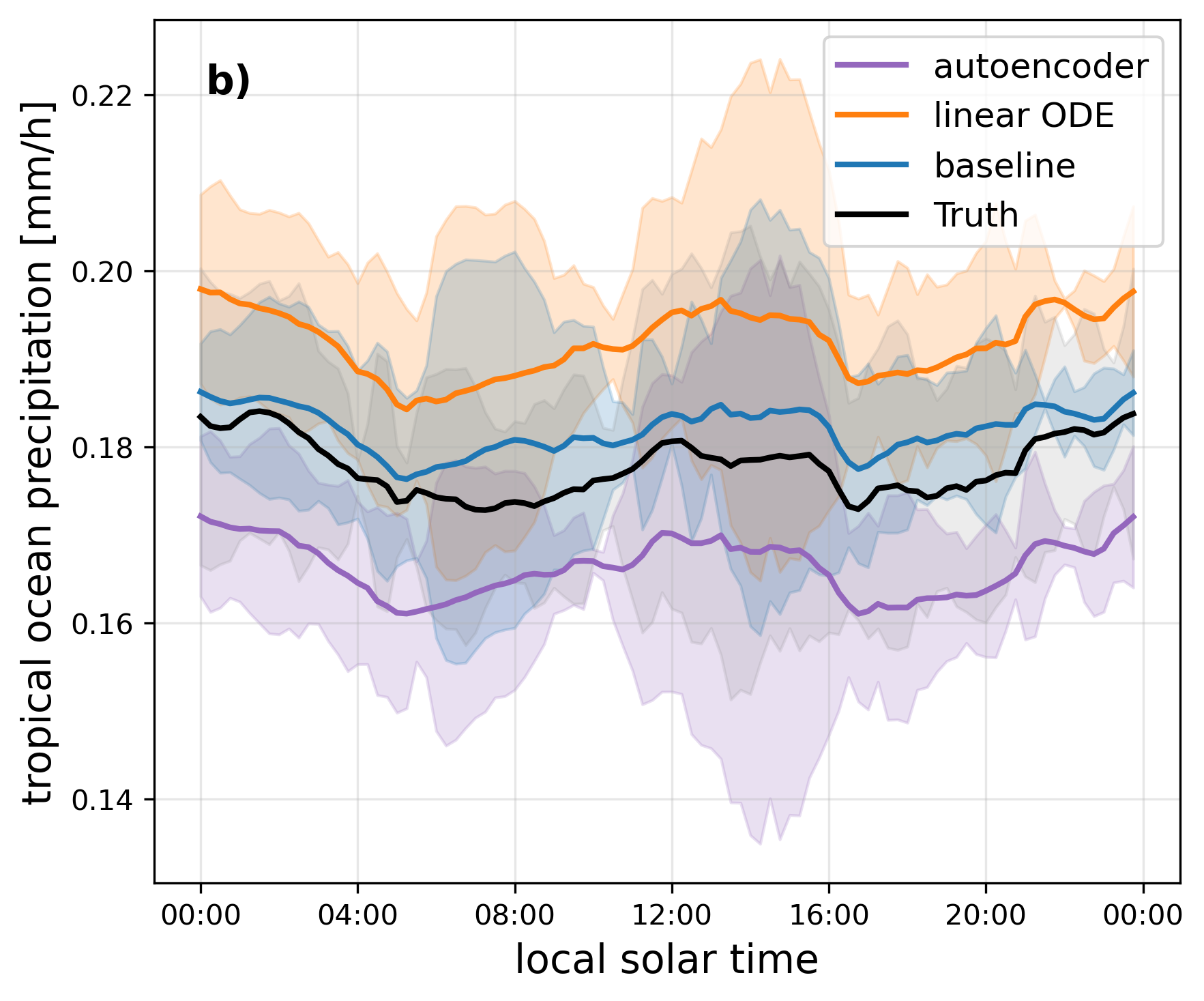}
    \end{minipage}
    \caption{Prognostic memory helps capture the evening and morning precipitation peaks over tropical land, in particular after symbolic distillation to a linear ODE (a), but yields no clear improvement over tropical ocean (b). Shading denotes 95\% confidence intervals.}
    \label{fig:DYAMOND_temporal_patterns}
\end{figure}

Next we construct mean precipitation diurnal cycles as a function of longitude-specific local solar time (LST), averaged over tropical land and ocean.  %
The peak time $\varphi$ and intensity $A$ are defined from these mean cycles based on the time of maximum precipitation rate. Over tropical land, the memory-informed parameterizations reproduce the temporal evolution of the reference simulation remarkably well (Figure \ref{fig:DYAMOND_temporal_patterns}a). The linear ODE and autoencoder models produce nearly identical precipitation cycles, with only minor differences near the midday minimum. The peak timing and intensity over tropical land generally agree with the reference values of approximately $\varphi = 18{:}30$ LST and $A = 0.21,\mathrm{mm,h^{-1}}$. The simulated peak timing is also consistent with observational estimates that place the diurnal cycle phase near 18:00 LST \cite{christopoulos_assessing_2021}. The DYAMOND simulation also exhibits a weaker secondary morning peak, which may reflect the persistence or propagation of organized convective systems. The memory-informed parameterizations also reproduce this bimodal structure faithfully. In contrast the baseline parameterization %
appears to overestimate the instantaneous connection between solar heating and rainfall, producing a precipitation maximum near $\varphi_{\text{baseline}}=12{:}00$ LST, when the reference simulation instead exhibits a local minimum. This behavior is common in ESMs with parameterizations lacking memory as most CMIP models \cite{christopoulos_assessing_2021}. While solar heating plays a key role in initiating convection, precipitation typically occurs only after convective systems have had sufficient time to develop and organize. By incorporating information from previous time steps, the memory-informed parameterizations are able to represent this delayed response more realistically.

Over tropical oceans, precipitation exhibits substantially weaker diurnal variability (Figure \ref{fig:DYAMOND_temporal_patterns}). Consequently, the estimated peak timing is sensitive to relatively small differences in precipitation intensity. Both the DYAMOND reference simulation and the learned parameterizations show precipitation maxima near midnight, whereas observational estimates place the phase closer to 06:00 LST \cite{christopoulos_assessing_2021}. %
The memory-informed parameterizations reproduce the limited temporal variability of the reference simulation but exhibit a constant precipitation bias towards opposite directions. In contrast, the baseline parameterization shows a smaller mean bias.

Overall, the analysis suggests that incorporating memory substantially improves the representation of precipitation timing over tropical land. In contrast, the benefits over tropical oceans are less clear. While the memory-informed parameterizations reproduce the weak temporal variability present in the reference simulation, they also exhibit a persistent precipitation bias that is observed to a lesser extent in the baseline model. One possible explanation is that the memory models preferentially learn temporal relationships associated with the pronounced land diurnal cycle and apply similar dynamics in regions where the temporal signal is considerably weaker. In such a scenario, the temporal dependencies learned over land may not generalize optimally to oceanic conditions, resulting in a systematic bias rather than an improved representation of the comparatively weak oceanic cycle. %

\subsection{Symbolic Prognostic Memory Variables of Convection} \label{sec:ODE_interpretation_DYAMOND}
In agreement with the L96 experiment, prognostic memory variables are derived via a linear ODE such that we can perform a similar analysis. In the precipitation case, the latent space is forced by multiple physical variables, enabling analysis of the $\mathbf{\Xi}_x$ matrix. Since all forcing variables are standardized, the coefficients quantify the sensitivity of the latent evolution to specific physical anomalies. %
Each row $\Xi_b + \Xi_x\mathbf{x}$ in \eqref{eq:linear_dZdt_solution} only impacts its corresponding latent space variable, so we can analyze each row independently to understand which combinations of physical variables the latent space finds informative. %

In general, temperature-related variables like $T_{\text{2m}}, T_{\text{sfc}}$ and $H$ form the dominant contributors to the equations (Figure \ref{fig:Xi_L96_DYAMOND}). Moisture-related variables have smaller coefficients and each equation only picks either $q_{\text{2m}}$ or $PW$ with the exception of $\dot z_2$. Latent surface heat fluxes are relatively small across all equations, with a minor contribution to $\dot z_1$. In equation $\dot z_2$ we obtain a 2m moisture anomaly with respect to column water vapor balanced by near-surface temperature. This balance could reflect the thermodynamic triggering of moist convection, where the interplay between low-level moisture availability and surface heating determines whether the boundary layer overcomes convective inhibition to initiate buoyant updrafts. However, interpretation of patterns in the linear coefficients is difficult due to the standardized inputs. Yet, the linear ODE provides an improvement in terms of interpretability over the autoencoder which can be further investigated by looking at the timescales of the systems eigenmodes.

The eigenvalues $\lambda_i$ characterize the stability of the linear operator $\Xi_z$ as a whole, governing the decay of the system's eigenmodes rather than the raw latent variables $z_i$ directly. Since the self-interaction submatrix is not diagonal, the physical latent variables are mixtures of these eigenmodes. However, these eigenvalues still estimate the intrinsic timescales of the dynamical system, with half-lifes ranging from approximately 50 minutes to nearly 3 hours (see Figure \ref{fig:Xi_L96_DYAMOND}). The first two latent space dimensions describe an oscillating mode. However, the oscillation period of 98 hours is too large for the rotational dynamics to have significant effects given significantly smaller half-life times of $t^{(1,2)}_{1/2}=0.86$\,h.

With the longest half-life being $t^{(4)}_{1/2}=2.8$ hours, our memory-informed parameterization might improve over the baseline model by capturing the non-equilibrium response of convection subject to atmospheric perturbations \cite{davies_simple_2009}. This time scale is also consistent with findings from \citeA{colin_identifying_2019} who showed that, in the absence of mesoscale organization, convection will take a few hours to recover after a perturbation to the environmental fine structure.

A prognostic organization variable $org$ was postulated by \citeA{mapes_parameterizing_2011}, evolved by a linear ODE forced by the rain evaporation rate. The $org$ variable was used to perturb parameters of the convection scheme with the aim of improving precipitation estimates. While their setup shows some structural differences compared to ours, such as a three-dimensional $org$-variable, subject to a steering flow derived from the mass-weighted mean of the convective layer and different forcings, results agree in that a linear process can carry information on the unresolved state. While \citeA{mapes_parameterizing_2011} set their $org$-timescale loosely based on prior research to $3\,$h, one of our eigenmodes might cover similar dynamics with $t^{(4)}_{1/2}=2.8\,$h. In contrast to \citeA{mapes_parameterizing_2011}, our prognostic memory variables improve diurnal precipitation cycles.  

\section{Conclusion and Outlook} \label{sec:conclusion}
Incorporating past observables as input variables to capture implicit sub-grid state information has been widely discussed in recent research on ESM parameterizations \cite{han_ensemble_2023, lin_navigating_2025, heuer_beyond_2026, behrens_simulating_2025, beucler_distilling_2025} and conceptual models like L96 \cite{parthipan_using_2023, bhouri_memory-based_2023, brolly_stochastic_2025}. In contrast, this work explicitly encodes sub-grid state information by expanding the prognostic variable set through symbolic distillation of an autoencoder, which compresses information from past observables into a low-dimensional latent space.

Our approach is tested through two experiments: an online Lorenz-96 testbed and an offline coarse-scale precipitation parameterization trained on high-resolution data. In the L96 experiment the additional prognostic information yields improved climate, weather, and regime statistics, while in the precipitation experiment they lead to accurate reproduction of the diurnal cycle over tropical land and spatial patterns of tropical precipitation. Notably, the symbolically distilled prognostic memory variables via a linear ODE achieve many of the benefits of a complete inclusion, for example substantially improving the diurnal cycle of tropical land precipitation, demonstrating that compact, physically interpretable prognostics can capture essential temporal processes absent in traditional diagnostic parameterizations.

The proposed framework provides a pathway to integrate useful information into ESMs without inflating the input space with multiple past time steps of high-dimensional variables, which would increase model complexity and does not lead to physical interpretability. Because our prognostics are data-driven, given suitable training data, they can operate across multiple timescales and represent any desired combinations of physical processes simultaneously. We consider this work a proof-of-concept, as a true prognostic convection parameterization in an ESM must address additional challenges. Initialization of the prognostic variables in the ESM would be straightforward, as our results show that the linear ODE quickly becomes independent of its initial conditions. However, so far, our applications are restricted to one- (L96) and two-dimensional (coarse-grained precipitation) cases, meaning future research should explore how to generate three-dimensional prognostics for ESMs. A three-dimensional prognostic variable representing organized convection should respond to dynamical processes like advection \cite{mapes_parameterizing_2011}, which plays a key role in the transport and organization of convective systems. In the current framework, advection is not incorporated, and future training strategies should account for this by embedding advection-aware dynamics into the prognostic equations. This will be crucial for ensuring that the prognostic variables evolve realistically under large-scale flow and maintain their predictive skill over time.

Our result advance that of \citeA{shamekh_implicit_2023}, showing that convective organization metrics can be derived from high-resolution precipitable water fields to improve coarse-grained precipitation predictions. However, lacking a stable evolution of these metrics over multi-step roll-outs when using only coarse-scale variables. In contrast to this study, we demonstrate that our linear ODE governing the additional prognostics is numerically stable and retains information effectively during long roll-outs. As a limiting factor, our online experiments are restricted to the L96 system, and it remains an open question whether these prognostic variables also stabilize the parameterization itself in more complex configurations. 

Integrating prognostic variables into ML parameterizations represents a logical next step in hybrid ESM development, mirroring the evolution of their conventional counterparts \cite{cohen_unified_2020, thayer-calder_unified_2015}. Our framework outlines a general approach that can accommodate prognostics governed by nonlinear ODEs. However, linear ODEs proved capable of replicating the latent dynamics and offered greater flexibility when adapted during roll-out training alongside the parameterization. Without the need for a general right-hand side formulation, the linear ODE can be learned using more tailored ML techniques. Future research should continue to explore how prognostic variables for ML parameterizations could be learned through Koopman Autoencoders (KAEs) \cite{lusch_deep_2018}, as previously demonstrated for moment-based microphysics schemes \cite{lamb_reduced-order_2024}. This framework enables fast end-to-end GPU training and could yield even better results, particularly with more complex autoencoder architectures like three-dimensional inputs. However, future improvements in equation discovery algorithms may justify revisiting prognostics governed by nonlinear ODEs, warranting further developments in nonlinear symbolic distillation.

\newpage

\appendix
\section{Hyperparameters}\label{app:hyper_opt}
Table \ref{tab:hyperparameters} summarizes hyperparameters of our ML-architecture, as well as tuned hyperparameters marked with $^*$. The number of past time steps used ($M$) was tuned outside the main hyperparameter optimization procedure solely focusing on parameterization performance (see \ref{tab:hyper_opt_M}).\\
The DYAMOND-precipitation encoder uses a one-dimensional convolution applied to $\mathbf{X}_{\text{past}}$ of shape $(N_m, M)$, where $N_m$ is the number of memory variables and $M$ is the number of past time steps. The convolution is performed along the temporal dimension $M$, enabling the extraction of local temporal features across the memory variables. The number of convolutional output channels is scaled with the input size, $C_{\mathrm{out}} \propto N_m \cdot M$, to ensure sufficient representational capacity prior to compression into the latent space.\\
We construct $\mathbf{X}_{\text{past}}$ on-the-fly during the forward pass by sampling sequences of length $N_{T_{\text{batch}}}$ from the dataset and forming time-delayed embeddings directly on the GPU. This avoids explicitly storing $M$ shifted copies of the input, reducing memory usage by approximately a factor of $M$ and enabling the full training dataset to fit into GPU memory.
\begin{table}[]
    \centering
    \begin{tabular}{l|l|l}
    Parameter & L96 & DYAMOND precipitation \\ \hline
    Past time steps $M^*$ & 1000 & 20 \\
    Latent space dimensions $d_z^*$ & 6 & 4 \\
    Weight decay $w^*$ & 0.0 & $10^{-6}$ \\
    Nodes per layer (NN) & 16 & 16 \\
    Number of layers (NN) & 6 & 5 \\
    $\alpha$ & 0.5 & 0.5 \\
    Memory features $N_m$ & 1 & 6 \\
    Kernel size $k$ & -- & 3 \\
    Conv. output channels $C_{\mathrm{out}}$ & -- & $M \cdot N_m = 120$ \\
    Encoder structure 
    & $[M, 32, 16, 12, d_z]$
    & Conv1D($N_m \rightarrow C_{\mathrm{out}}$) $\rightarrow$ [32, 16, 12, $d_z$] \\
    Activation function & ReLU & ReLU \\
    Learning rate & $7 \cdot 10^{-3}$ & $10^{-3}$ \\
    $N_{T_{batch}}$ & 10.000 & 2137 \\
    Batch size & $8(N_{T_{batch}}-M)$ & $64(N_{T_{batch}}-M)$ \\
    Epochs & 100 & 150
    \end{tabular}
    \caption{Model architecture and training hyperparameters. Parameters included in structured hyperparameter optimization are marked with $^*$. Convolutional layers are not used for L96. Therefore, no kernel size or convolutional output channels are listed.}
    \label{tab:hyperparameters}
\end{table}

\begin{table}[]
    \centering
    \caption{Setting the number of past time steps used for training}
    \begin{tabular}{lll}
    past time steps $M$ & $d_z$ & $R^2$ validation set \\ \hline
    L96 \\
    10 & 4 & 0.882 \\
    100 & 4 & 0.919 \\
    1000 & 4 & 0.931 \\
    10 & 8 & 0.882 \\
    100 & 8 & 0.919 \\
    1000 & 8 & \textbf{0.937} \\
    \hline
    DYAMOND precipitation \\ 
    3 & 4 & 0.644 \\
    5 & 4 & 0.674 \\
    10 & 4 & 0.686 \\
    20 & 4 & \textbf{0.691} \\
    3 & 8 & 0.648 \\
    5 & 8 & 0.674 \\
    10 & 8 & 0.690 \\
    20 & 8 & 0.689 \\ 
    \label{tab:hyper_opt_M}
    \end{tabular}
\end{table}

\newpage

\section*{Open Research Section}
The code used in this study is openly available on GitHub at https://github.com/DLR-PA-EVA/schoenfeld26james\_PrognosticMemoryParmeterizations. The DYAMOND data can be accessed through the archive of the German Climate Computing Center (DKRZ) which managed the data under the ESiWACE and ESiWACE2 projects.

\section*{Conflict of Interest disclosure}
The authors declare there are no conflicts of interest for this manuscript.

\acknowledgments
Jurij Sch\"onfeld, Julien Savre and Veronika Eyring received funding for this study from the European Research Council (ERC) Synergy Grant “Understanding and Modelling the EarthSystem with Machine Learning (USMILE)” under the Horizon 2020 research and innovation programme (Grant agreement No. 855187). Veronika Eyring was additionally supported by the Deutsche Forschungsgemeinschaft (DFG, German Research Foundation) through the Gottfried Wilhelm Leibniz Prize awarded to Veronika Eyring (Reference No. EY 22/2-1). Jurij Sch\"onfeld acknowledges additional support from the EERIE project (grant agreement no. 101081383) funded by the European Union. This work received funding from the EU Horizon Europe project “Artificial Intelligence for enhanced representation of processes and extremes in Earth System Models (AI4PEX)” (Grant agreement ID: 101137682). Tom Beucler received support from AIPEX, funded by the Swiss State Secretariat for Education, Research and Innovation (SERI, Grant 23.00546). Sherwood received funding from the Australian Research Council CE230100012.  This work used resources of the Deutsches Klimarechenzentrum (DKRZ) granted by its Scientific Steering Committee (WLA) under project ID bd1179 and bb1153 to access the DYAMOND winter data.

\bibliography{references}

\end{document}